# Minimum Error Tree Decomposition

**Lifu Liu** and **David C. Wilkins**
Department of Computer Science
University of Illinois
405 North Mathews Ave
Urbana, IL 61801

**Xiaoxin Ying** and **Zhaoqi Bian**
Department of Automation
Tsinghua University
Beijing,100084,P.R.China

## Abstract

This paper describes a generalization of previous methods for constructing tree-structured belief network with hidden variables. The major new feature of the described method is the ability to produce a tree decomposition even when there are errors in the correlation data among the input variables. This is an important extension of existing methods since the correlational coefficients usually cannot be measured with precision. The technique involves using a greedy search algorithm that locally minimizes an error function.

## I. Introduction

In belief network reasoning and learning causal structure, it is desired to decompose distribution $P(x_1,\cdots,x_n)$ of $n$ binary stochastic variables into marginals of $n+1$ variables $P(x_1,\cdots,x_n,w)$, in order to render $x_1,\cdots,x_n$ conditionally independent given $w$ (J. Pearl 1986).

This decomposition problem is very difficult to solve directly (P.Lazarfeld 1966). Pearl's solution approach to the problem was to replace the hidden variable $w$ with a group of hidden variables $w_1,\cdots,w_m, 1 \leq m \leq n-2$, and then construct a tree that has $n$ leaves corresponding to the stochastic variables $x_1,\cdots,x_n$ and has $m$ internal hidden nodes $w_1,\cdots,w_m$ (Pearl, 1986). Thus the $P(x_1,\cdots,x_n)$ is represented as the margin of the correlation distribution of the nodes in the tree,

$$P(x_1,\cdots,x_n,w_1,\cdots,w_m), 1 \leq m \leq n-2.$$

The first step of Pearl's method is to define the four allowable topologies of quadruplets, where each quadruplet has four leaves $i, j, k, l$, and two internal nodes $w_1$ and $w_2$, and each of the internal nodes is connected to a pair of the leaves. Given this fundamental tree substructure, Pearl's method defines a sufficient condition for connecting four leaves into a quadruplet. This sufficient condition is then applied repeatedly to construct a tree.

In this paper, we extend the excellent approach to Pearl by defining a weaker sufficient condition for connecting four leaves into a quadruplet. This weaker sufficient condition is satisfied in situations where the input information is noisy or incomplete, which is typically the case in actual applications.

The sufficient condition for the topology of quadruplet used by Pearl is

$$\rho_{ik}\rho_{jl} = \rho_{il}\rho_{jk}, \qquad (1)$$

given that it is known that $P(x_1,\cdots,x_n)$ can be written as the marginal probability of an $(n+m)$ variable distribution $P_s(x_1,\cdots,x_n,w_1,\cdots,w_m)$ in which $x_1,\cdots,x_n$ are conditionally independent given $w_1,\cdots,w_n$, i.e.,

$$P_s(x_1,\cdots,x_n,w_1,\cdots,w_m) = \qquad (2)$$

$$\prod_{i=1}^{n} P_s(x_i|w_j) P_s(w_j|w_k) \cdots P(w_m) \qquad (3)$$

where

$$\rho_{ij} = \frac{(p_{ij} - p_i p_j)}{(p_i(1-p_i))^{1/2}(p_j(1-p_j))^{1/2}},$$

$$p_i = P(x_i = 1),$$

$$p_{ij} = P(x_i = 1, x_j = 1).$$

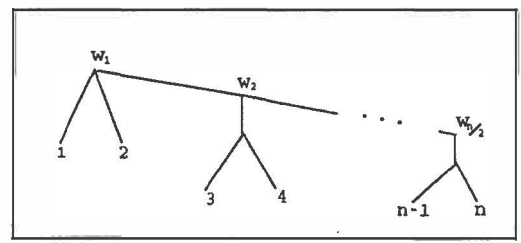

Figure 1. The type of tree structure that is produced by repeated application of equation (1).

An ideal decomposition will satisfy the conditions (1) for all nodes connected to internal nodes. Unfortunately, this is often not the case, and so the decomposition algorithm will terminate without producing a solution. An approach that circumvents this problem is to minimize the error $|\rho_{ik}\rho_{jl} - \rho_{il}\rho_{jk}|$.

So as to illustrate the challenges of constructing an approximate tree decomposition, we will briefly describe an earlier approach that we abandoned. Suppose the tree has the structure like in Figure 1 and the structure of the tree is characterized by equations

$$\rho_{13}\rho_{24} = \rho_{14}\rho_{23}$$

$$\rho_{15}\rho_{26} = \rho_{16}\rho_{25}$$

$$\vdots$$

$$\rho_{1,n}\rho_{2n} = \rho_{1n}\rho_{2,n-1} \quad (4)$$

$$\vdots$$

$$\rho_{35}\rho_{46} = \rho_{36}\rho_{45}$$

$$\vdots$$

$$\rho_{3-n,n-1}\rho_{n-2,n} = \rho_{n-3,n}\rho_{n-2,n-1}$$

Given this information, one possible approach to construct the tree in Figure 1 is

1. Compute all possible $\rho_{ij}, 1 \leq i \leq n, 1 \leq j \leq n, i \neq j$.

2. Construct a search tree as shown in Figure 2. During the construction, calculate

$$\sum |\rho_{i,k}\rho_{j,l} - \rho_{j,k}\rho_{i,l}| +$$

$$\sum |\rho_{i',k'}\rho_{j',l'} - \rho_{j',k'}\rho_{i',l'}| + \ldots$$

as the current minimum error for all nodes. If a node tested has a sum greater than the current minimum error, then discard the node and all its subsequent nodes, and go on testing its right neighbor node.

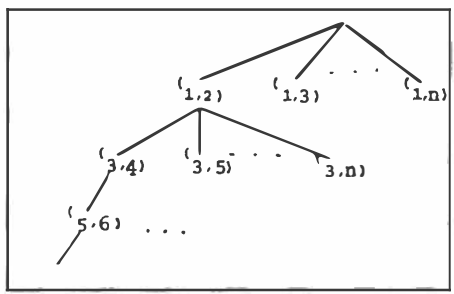

Figure 2. The search tree that models a hypothetical approach to approximate tree decomposition. In this approach, the search is abandoned along tree branches below a node if the error at the node is greater than the current minimum error.

It can be easily deduced that this procedure requires exponential time, and needs an additional step to determine the connecting topology once the node quadruplets are determined. This ends our presentation of our earlier approach, which was presented to illustrate some of the problems that an efficient method for approximate tree decomposition must confront. The remainder of the paper presents our computationally tractable approach to approximate tree decomposition.

Let us begin our approach by presenting the following two observations. The first observation is that in a tree structured causal model there should be corresponding links among correlations for each pair of nodes. For example, we have the following equations

$$\rho_{15}\rho_{26} = \rho_{16}\rho_{25}$$

$$\rho_{13}\rho_{24} = \rho_{23}\rho_{14}$$

$$\rho_{36}\rho_{45} = \rho_{46}\rho_{35}$$

for the six variables in Figure 2.

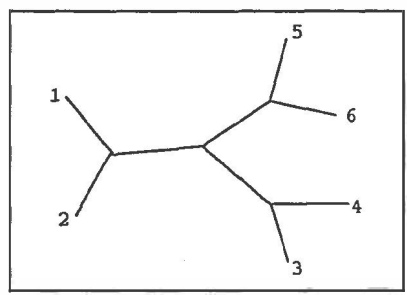

Figure 3. A tree structure for six variables.

Now these equations sometimes may not hold due to errors in estimations of $\rho_{ij}$, even though the variables themselves are tree-decomposable.

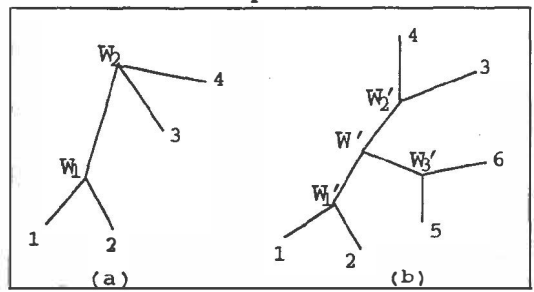

Figure 4. Reliability of correlations. The pairs (1, 2) and (3, 4) in (a) are connected via two internal nodes (reasons), therefore their correlations are more "reliable" than the ones between pairs (1, 2) and (3, 4) in (b).

Our second main observation is that some correlations are more likely in a decomposed tree than the others. Examine two trees in Figure 4; correlations between pairs (1,2) and (3,4) in (a) are thought to be more "reliable" than that in (b). From a mathematical point of view, correlations in (a)

$$\rho_{14} = \rho_{1w_1}\rho_{w_1w_2}\rho_{w_24}$$

$$\rho_{24} = \rho_{2w_1}\rho_{w_1w_2}\rho_{w_24}$$

$$\rho_{13} = \rho_{1w_1}\rho_{w_1w_2}\rho_{w_23}$$

$$\rho_{23} = \rho_{2w_1}\rho_{w_1w_2}\rho_{w_23}$$

are affected by 9 parameters, while in (b), these values

$$\rho_{14} = \rho_{1w_1}\rho_{w_1w'}\rho_{w'w_2}\rho_{w_24}$$

$$\rho_{24} = \rho_{2w_1}\rho_{w_1w'}\rho_{w'w_2}\rho_{w_24}$$

$$\rho_{13} = \rho_{1w_1}\rho_{w_1w'}\rho_{w'w_2}\rho_{w_23}$$

$$\rho_{23} = \rho_{2w_1}\rho_{w_1w'}\rho_{w'w_2}\rho_{w_23}$$

are subject to the influence of 10 parameters; two of them are also correlated with variables 5 and 6.

What we want is to find a step-by-step way to build a tree structured network by connecting small structures together, piece by piece. This cannot be done unless some rule of thumb be set first. So we make our assumptions from the above discussion. The first assumption is that there should be a criterion to justify connecting two tree structures, which has something to do with correlations of variables within the structures and reflects the connectability of the two structures. The value, or error, $|\rho_{ik}\rho_{jl} - \rho_{jk}\rho_{il}|$ serves the needs quite well. If it remains small for all possible $i, j, k, l$ in the two structures and if the connected structure is going to connect another one then a smaller error can be expected. On the other hand, the average error usually will rise as the internal nodes increase. The details of the definitions are coming up next.

## II. Errors Between Trees and Nodes

Let us look at the structure of a decomposed tree.

### Definition 1: Tree Structure

The bottom nodes of a tree are all leaf nodes, i.e., the stochastic variables of a causal network. The top node of a tree is its root. There are a number of nonterminating nodes, that is, the intermediate nodes connected between the leaves and the root, with levels $0, 1, \cdots, l-1$ from the lowest to the highest. There leaf nodes are referred to as nodes belonging to tree $w$, noted by

$$w = \{i | 1 \leq i \leq n\}.$$

As shall later become apparent, the root is redundant in the final decomposed tree but is essential in the process because all combinations take place at roots. Figure 5 gives an illustration of a tree with 7 leaves and 5 nonterminating nodes.

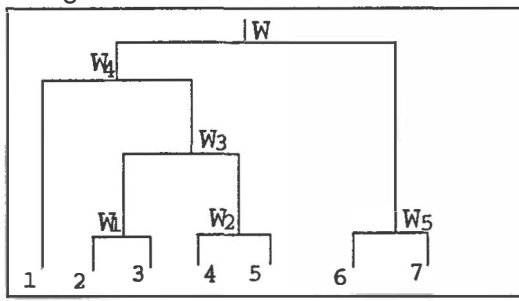

Figure 5. A tree with 7 leaves and 5 nonterminating nodes.

Leaves that do not belong to any trees are called the independent leaf nodes or the independent nodes.

Now let us consider the rules of combination of nodes and trees.

### Definition 2: Combination Operations

Two pairs of independent nodes $(i, j)$ and $(k, l)$ can be connected together to form a tree $w$. The root of the tree is $w$. See Figure 6.

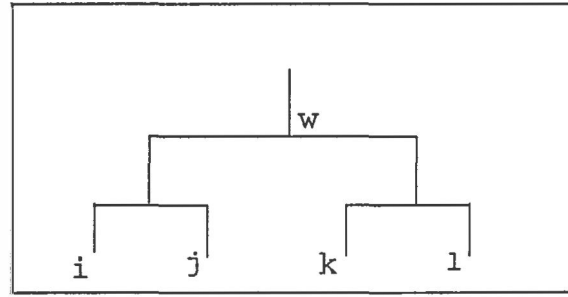

Figure 6. Combination of two independent node pairs.

A pair of nodes $(i, j)$ can be combined to a tree $w, w = \{1, \cdots, i\}$, to form a new tree $w'$. The root of the new tree is $w'$, and root $w$ becomes a nonterminating node. See Figure 7.

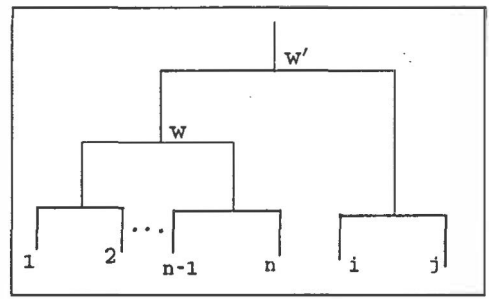

Figure 7. Combination of an independent node pair with a tree.

Two trees $w_1$ and $w_2$ can be combined to form a new tree $w$. The roots $w_1$ and $w_2$ become two nonterminating nodes as in Figure 8.

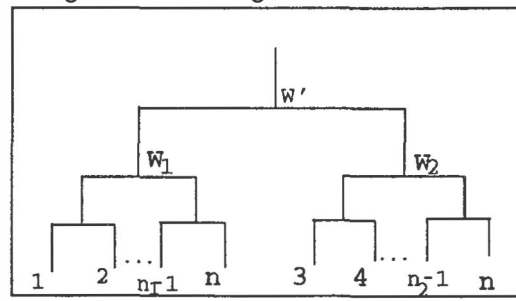

Figure 8. Combination of two trees.

For one independent node $i$ and a tree rooted at $w$, a new root $w'$ is formed, with two sons of node $i$ and the non-terminating node $w$, the original root. See Figure 9.

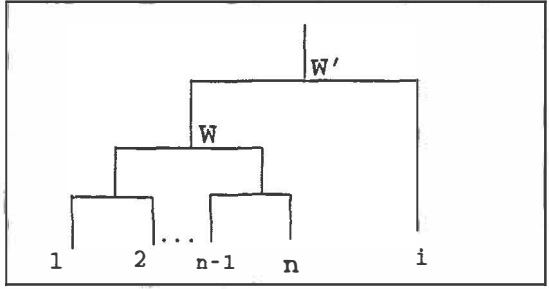

Figure 9. Combination of one independent node with a tree.

Given the above four operations for combining nodes and trees, many ways can be found to decompose a causal network, depending on how the nodes are grouped and connected. It should be noted that many of the networks cannot be precisely decomposed to a tree structured equivalent at all. Nevertheless, many approximations exist which provide a practical solution for network reasoning.

The necessary condition used by Pearl to allow a network with four logical variables $(i,j,k,l)$ to be decomposed into a tree is

$$|\rho_{ik}\rho_{jl} - \rho_{il}\rho_{jk}| = 0. \qquad (5)$$

If the expression does not yield zero and if we still want to decompose the network then an error will result; the value of (3) is a good measure of the error. It is possible to view this value as the error between the decomposed tree-structured network and the original one. Let us define four such errors.

**Definition 3: Decomposition Errors**

Let $\rho_{xy}$ be the correlation coefficient of leaf nodes $x$ and $y$, then

1. the error between node pairs $(i,j)$ and $(k,l)$ is

$$d_{ij/kl} = |\rho_{ik}\rho_{jl} - \rho_{il}\rho_{jk}|; \qquad (6)$$

2. the error between node pair $(i,j)$ and tree $w$ is

$$d_{ij/w} = d_{w/ij} = \max_{k,l \in w_{k \neq l}} |\rho_{ik}\rho_{jl} - \rho_{il}\rho_{jk}|; \qquad (7)$$

3. the error between two trees $w_1$ and $w_2$ is

$$d_{w_1/w_2} = d_{w_2/w_1} = \max_{i,j \in w_1, i \neq j, k, l \in w_2, k \neq l} |\rho_{ik}\rho_{jl}-\rho_{il}\rho_{jk}|; \qquad (8)$$

and the error between a single leaf node $i$ and a tree $w$ is

$$d_{i/w} = d_{w/i} = \max_{j,k,l \in w_{j \neq k \neq l}} |\rho_{ik}\rho_{jl} - \rho_{il}\rho_{jk}|. \qquad (9)$$

## III. The Minimum Error Tree Decomposition Algorithm

**Algorithm Stage 1:**
**Determination of Overall Tree Topology**

Suppose initially there are N independent leaf nodes. At all points during the tree construction, use the definitions of errors presented in the previous section of this paper to calculate the errors (1) between two independent node pairs, (2) between an independent pair and a tree, (3) between two trees, and (4) between a single independent node and a tree. Having found the nodes and trees that have a minimum error, combine them according to the operations defined earlier.

Four important points or elaborations of our algorithm are as follows.

1. If $N = 3$, i.e., the number of all leaf nodes is three, a tree structured equivalent network is simply the star-decomposition of the three nodes (J. Pearl 1986).

2. If several identical errors are found in the process of tree decomposition, they can be combined one by one in any sequence. To limit uncertainty of the final tree structure, constraints may be applied. For example, set the precedence of combination as tree to tree, tree to node pair, node pair to node pair, and tree to single node. Another method is to compare the errors $d_{ij/kl}$ for all possible $(i,j,k,l)$

$$\min_{i,j,k,l} |\rho_{ik}\rho_{jl} - \rho_{il}\rho_{jk}|,$$

and combine the corresponding nodes and/or trees first, then repeat the process.

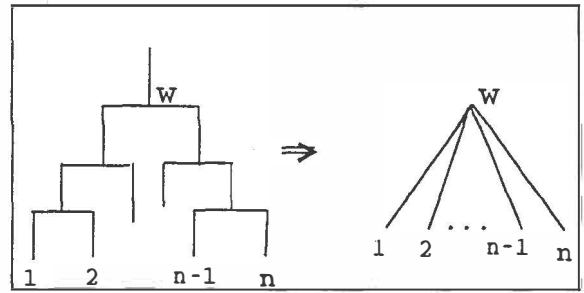

Figure 10. Tree structure simplification.

3. The tree-tree and tree-node-pair errors, beside the $\max_{i,j,k,l}|\rho_{ik}\rho_{jl} - \rho_{il}\rho_{jk}|$, are defined as

$$\frac{\sum_{i,j,k,l}|\rho_{ik}\rho_{jl} - \rho_{il}\rho_{jk}|}{\sum_{i,j,k,l} 1}. \qquad (10)$$

Here the denominator is the number of all possible combination of nodes $(i,j,k,l)$. However, they should not be defined as

$$\min_{i,j,k,l} |\rho_{ik}\rho_{jl} - \rho_{il}\rho_{jk}|,$$

because then the upper bound of the errors is unbounded.

4. The structure of the whole tree or any of its sub-trees can be simplified. This is done by removing all the non-terminating nodes between the root and the leaf nodes of the tree or sub-tree, and connecting all the leaf nodes to the root, as shown in Figure 10.

**Complexity of Algorithm Stage 1**

The complexity of the tree decomposition algorithm will now be considered. Note that the tree is a binary tree which may be restructured in the following form (Figure 11), which has the same number of non-terminating nodes as the original one. This number will not exceed the number of leaf nodes ($N-1$) in any case. To get every non-terminating node, at most

$$\binom{N}{4}$$

comparisons are needed. Thus the total number of comparisons is less than $O(N^5)$. All $|\rho_{ik}\rho_{jl} - \rho_{il}\rho_{jk}|$ need to be calculated once, and its computing complexity is $O(N^5)$.

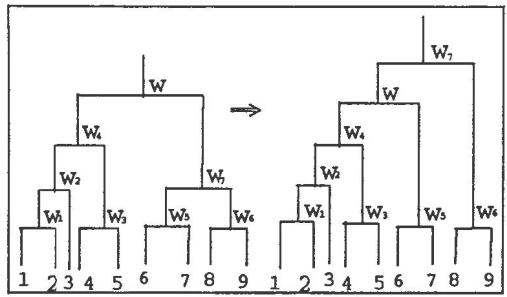

Figure 11. Tree reconstruction for complexity estimation.

**Decomposition Example**

In this example, we begin with a causal network of 15 variables. Let the initial set of independent leaf nodes be denoted by the set $I$, where

$$I = \{1, \cdots, 15\}.$$

Step 1:
Find the minimum error of quadruple

$$\min_{i,j,k,l \in I_{i \neq j \neq k \neq l}} d_{il/jk}.$$

In the example,

$$(i, j, k, l) = (5, 11, 12, 13).$$

Construct tree $w_1$, and we have

$$w_1 = \{5, 11, 12, 13\},$$

$$I = \{1, \cdots, 4, 6, \cdots, 10, 14, 15\}.$$

Step 2:
For $i, j \in w_1, k, l \in I$, and $i, j, k, l \in I$, and $i \in I, i, j, k, l \in w_1$, find the minimum error. We get

$$(i, j, k, l) = (3, 4, 9, 10).$$

Construct tree $w_2$, so

$$w_1 = unchanged,$$
$$w_2 = \{3, 4, 9, 10\},$$
$$I = \{1, 2, 6, 7, 8, 14, 15\}.$$

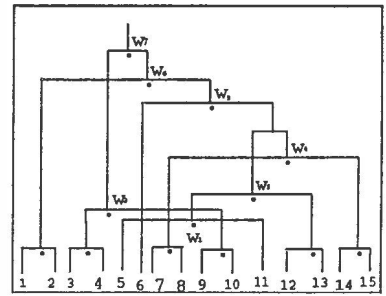

Figure 12. Tree construction example.

Step 3:
Following the same procedure, we found that $d_{7,8/14,15}$ to be the next minimum error. Thus

$$w_1 = w_2 = unchanged,$$
$$w_3 = \{7, 8, 14, 15\},$$
$$I = \{1, 2, 6\}.$$

Step 4:
Similarly, we get the minimum error $d_{w_1 w_2} = d_{11,13/7,15}$. Then we have

$$w_1 = w_2 = w_3 = unchanged,$$
$$w_4 = \{5, 7, 8, 11, \cdots, 15\},$$
$$I = unchanged.$$

Step 5:
Now the minimum error is $d_{6/w_4} = d_{6,7/12,13}$. Combine node 6 with tree $w_4$. Thus

$$w_1 = w_2 = w_3 = w_4 = unchanged,$$
$$w_5 = \{3, \cdots, 15\},$$
$$I = \{1, 2\}.$$

Step 6:
The minimum error is $d_{1,2/6,7}$. We construct tree $w_6$:

$$w_1 = w_2 = w_3 = w_4 = w_5 = unchanged,$$
$$w_6 = \{1, \cdots, 15\},$$
$$I = empty.$$

**Step 7:**
There is only one thing to do now: to combine trees $w_2$ and $w_6$. We get the final decomposed tree $w_7$ as in Figure 12.

**Algorithm Stage 2:**
**Calculation of Node Parameters**

Stage 1 of our minimum error tree decomposition algorithm determines the topological structure of the decomposed tree. The next stage of our algorithm is now presented, that determines the parameters of the nodes, the relationship among the internal and leaf nodes, and the relationship among the internal nodes themselves. To be more specific, the objective of Stage 2 is to determine the correlations of the $m$ internal nodes $w_1, \cdots, w_m$, and their prior probabilities $p(w_1), \cdots, p(w_m), 0 \leq m \leq n-2$.

Now

$$\rho_{ij} = \rho_{iw_1} \cdots \rho_{w_m j}, \tag{11}$$

where $w_1, \cdots, w_m$ are the internal nodes lying on the path from leaf $i$ to leaf $j$, $i = 1, \cdots, n, j = 1, \cdots, n$, and $i \neq j$.

By letting

$$x_{ij} = \log \rho_{ij}, \tag{12}$$

equation (10) becomes

$$x_{ij} = x_{iw_1} + \cdots + x_{w_m j}, \tag{13}$$

where $i = 1, \cdots, n, j = 1, \cdots, n$, and $i \neq j$.

Now let

$$X_w = [x_{1w_1}, \cdots, x_{1w_m}, x_{2w_1}, \cdots, x_{2w_m}, \cdots, x_{nw_m}]^T$$

$$X_{ij} = [x_{12}, \cdots, x_{1n}, x_{23}, \cdots, x_{2n}, \cdots, x_{n-1,n}]^T$$

Therefore, equation (12) can be rewritten in matrix notations as:

$$A X_w = X_{ij} \tag{14}$$

in which $A$ is a coefficient matrix whose elements are either 1 or 0.

Since the number of internal nodes $m$ is always no greater than $n-2$, where $n$ is the number of leaf nodes, there will always be more equations in (13) than unknown variables. Solving (13) in the least square error sense, we get

$$\hat{X}_w = (A^T A)^{-1} A^T X_{ij} \tag{15}$$

as an estimation of $X_w$.

The equation used to solve $p(w_1), \cdots, p(w_m)$ and all $p(i|w_j)$ is

$$\rho_{iw_j} = \frac{(p(i|w_j) - p(i))p(w_j)}{[p(i)(1-p(i))]^{1/2}[p(w_j)(1-p(w_j))]^{1/2}}, \tag{16}$$

with the constraints

$$0 \leq p(i|w_j) \leq 1,$$
$$0 \leq p(w_j) \leq 1,$$

$1 \leq i \leq n, 1 \leq j \leq m$. Non-linear equations (15) can be solved using non-linear programming (see, for example, Luenberger). This method makes the square error of the following equations

$$\left( \frac{(\hat{p}(i|w_j) - p(i))\hat{p}(w_j)}{[p(i)(1-p(i))]^{1/2}[\hat{p}(w_j)(1-\hat{p}(w_j))]^{1/2}} - \rho_{iw_j} \right)^{1/2} \tag{17}$$

minimum under the constraints

$$0 \leq \hat{p}(i|w_j) \leq 1,$$
$$0 \leq \hat{p}(w_j) \leq 1,$$

$1 \leq i \leq n, 1 \leq j \leq m$. This yields the needed values of the internal node parameters $\hat{p}(w_j)$ and $\hat{p}(i|w_j)$.

## Acknowledgements

We express our great debt to the following members of the Knowledge-Base Systems Group at the University of Illinois for their very helpful comments on earlier drafts of this paper: Sheldon Nicholl, Yong Ma, Kenneth Roe and Mike Lingk. The comments of Judea Pearl and an anonymous reviewer were also very helpful. We also express our thanks to Professor Mary Terperly and Sharon Collins of the University of Illinois for editorial assistance. This research was supported in part by ONR grant N00014-88K0124 and an Arnold O. Beckman research award.